\documentclass[10pt,twocolumn,letterpaper]{article}

\usepackage{cvpr}              

\usepackage[dvipsnames]{xcolor}

\newif\ifdrafting 
\draftingtrue 
\ifdrafting
  \newcommand{\ds} [1] {\textcolor{red}{[DS: #1]}}
  
  \newcommand{\TODO}[1]{\textbf{\color{red}[TODO: #1]}}
  
    \newcommand{\red}[1]{{\color{red}#1}}
    \newcommand{\todo}[1]{{\color{red}#1}}

\else
  \newcommand{\ds} [1] {}  
  \newcommand{\TODO} [1] {}
  
  \newcommand{\red} [1] {}  
  \newcommand{\todo} [1] {}    
\fi

\definecolor{cvprblue}{rgb}{0.21,0.49,0.74}
\usepackage[pagebackref,breaklinks,colorlinks,citecolor=cvprblue]{hyperref}

\def\paperID{16480} 
\def\confName{CVPR}
\def\confYear{2024}
\def\modelname{ZeroNVS}
\usepackage[accsupp]{axessibility}

\title{ZeroNVS: Zero-Shot 360-Degree View 
Synthesis from a Single Image}

\author{
Kyle Sargent\textsuperscript{1},
Zizhang Li\textsuperscript{1},
Tanmay Shah\textsuperscript{2},
Charles Herrmann\textsuperscript{2},
Hong-Xing Yu\textsuperscript{1},
\\[0.2em]
Yunzhi Zhang\textsuperscript{1}, 
Eric Ryan Chan\textsuperscript{1},
Dmitry Lagun\textsuperscript{2},
Li Fei-Fei\textsuperscript{1},
Deqing Sun\textsuperscript{2},
Jiajun Wu\textsuperscript{1}
\\[0.5em]
\textsuperscript{1}Stanford University,
\textsuperscript{2}Google Research
}

\begin{document}
\maketitle

\begin{abstract}
We introduce a 3D-aware diffusion model, \modelname, for single-image novel view synthesis for in-the-wild scenes. While existing methods are designed for single objects with masked backgrounds, we propose new techniques to address challenges introduced by 
in-the-wild multi-object scenes with complex backgrounds. Specifically, we train a generative prior on a mixture of data sources that capture object-centric, indoor, and outdoor scenes.  To address issues from data mixture such as depth-scale ambiguity, we propose a novel camera conditioning parameterization and normalization scheme. Further, we observe that Score Distillation Sampling (SDS) tends to truncate the distribution of complex backgrounds during distillation of 360-degree scenes, and propose ``SDS anchoring'' to improve the diversity of synthesized novel views. Our model sets a new state-of-the-art result in LPIPS on the DTU dataset in the zero-shot setting, even outperforming methods specifically trained on DTU. We further adapt the challenging  Mip-NeRF 360 dataset as a new benchmark for single-image novel view synthesis, and demonstrate strong performance in this setting. Code and models are available at \href{https://kylesargent.github.io/zeronvs/}{this url}.
\end{abstract}
\vspace{-8mm}
\section{Introduction}
\begin{figure*}
    \small
        \newcommand{\Figsize}{0.12\linewidth}
        \newcommand{\betweenviews}{\hspace{-4mm}}
        \newcommand{\beforerow}{\hspace{-4mm}}
        \newcommand{\betweeninputandoutput}{\hspace{-2mm}}
        \begin{tabular}{cccccccc}
        
        \multicolumn{5}{l}{CO3D} \\[-1pt]
        \cmidrule(r){1-2}
        
        \beforerow 
        \includegraphics[width=\Figsize, height=\Figsize]{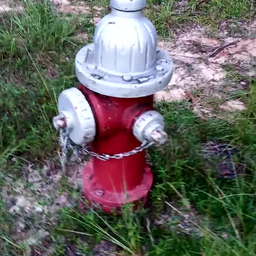} &
        \multicolumn{3}{c}{\betweeninputandoutput \includegraphics[height=\Figsize]{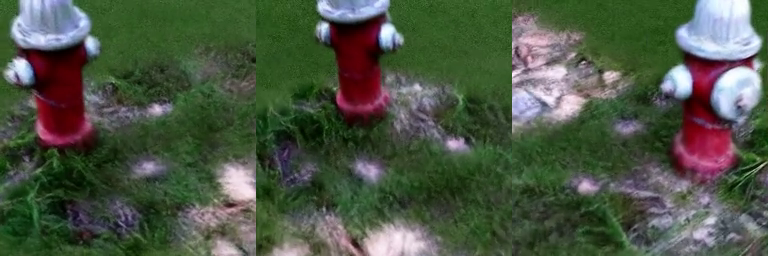}} & 
        
        \betweeninputandoutput \includegraphics[width=\Figsize, height=\Figsize]{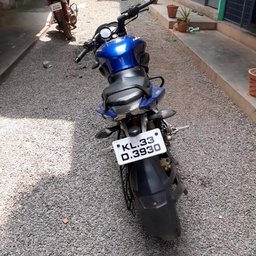} &
        \multicolumn{3}{c}{\betweeninputandoutput \includegraphics[height=\Figsize]{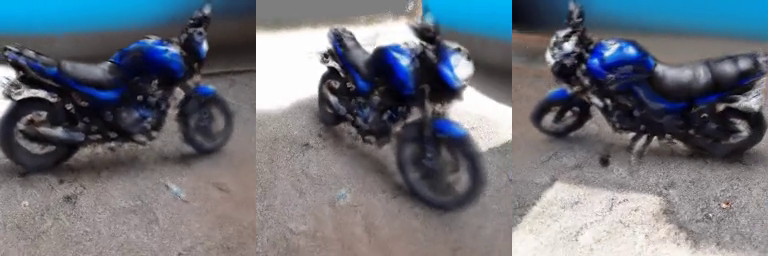}} \\[-1pt]

        \beforerow 
        \includegraphics[width=\Figsize, height=\Figsize]{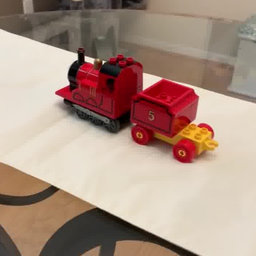} &
        \multicolumn{3}{c}{\betweeninputandoutput \includegraphics[height=\Figsize]{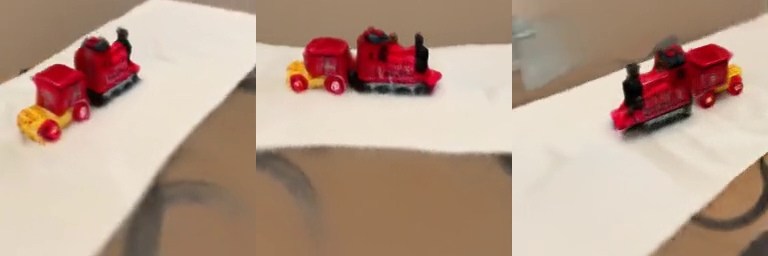}} & 
        
        \betweeninputandoutput \includegraphics[width=\Figsize, height=\Figsize]{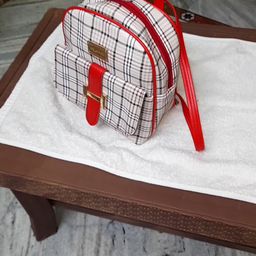} &
        \multicolumn{3}{c}{\betweeninputandoutput \includegraphics[height=\Figsize]{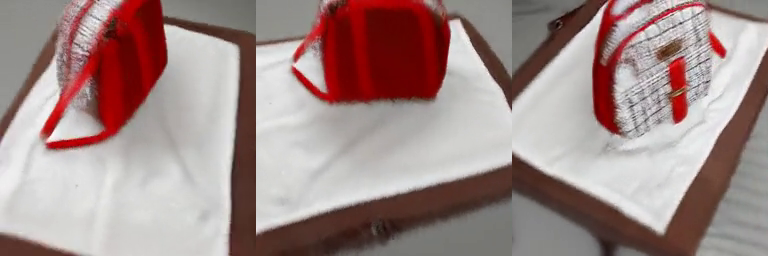}} \\[-1pt]
        
        \beforerow Input view & \multicolumn{3}{c}{\betweeninputandoutput ------------------------ Novel views ------------------------} & 
        Input view & \multicolumn{3}{c}{\betweeninputandoutput ------------------------ Novel views ------------------------} \\[4pt]
                
        \multicolumn{5}{l}{RealEstate10K} \\[-1pt]
        \cmidrule(r){1-2}

        \beforerow 
        \includegraphics[width=\Figsize, height=\Figsize]{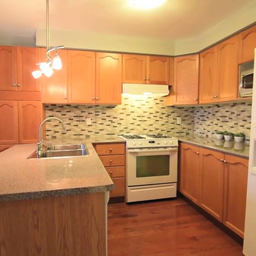} &
        \multicolumn{3}{c}{\betweeninputandoutput \includegraphics[height=\Figsize]{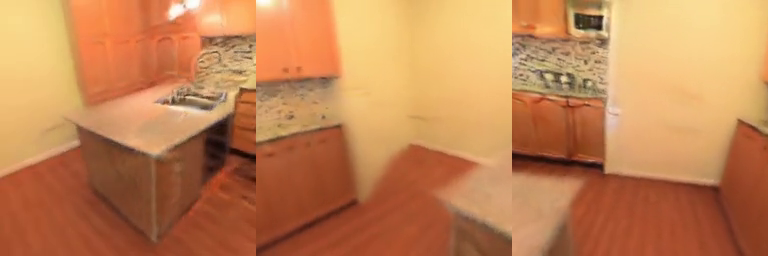}} & 
        
        \betweeninputandoutput \includegraphics[width=\Figsize, height=\Figsize]{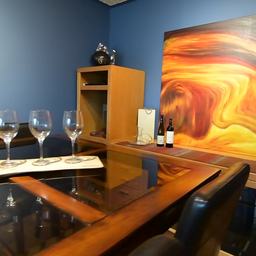} &
        \multicolumn{3}{c}{\betweeninputandoutput \includegraphics[height=\Figsize]{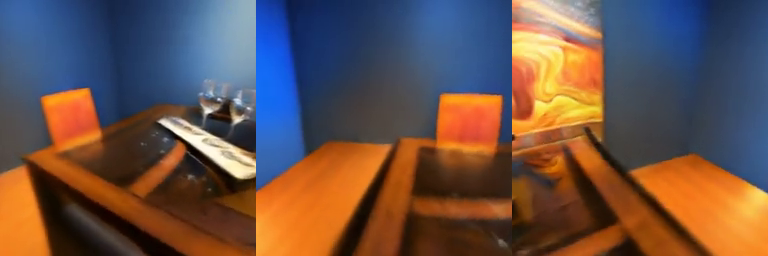}} \\[-1pt]

        \beforerow 
        \includegraphics[width=\Figsize, height=\Figsize]{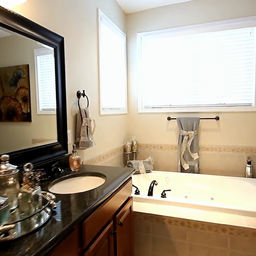} &
        \multicolumn{3}{c}{\betweeninputandoutput \includegraphics[height=\Figsize]{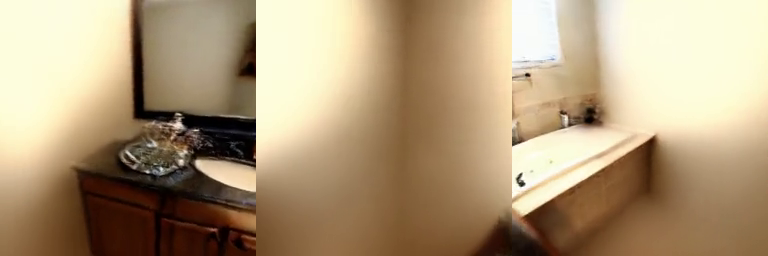}} & 
        
        \betweeninputandoutput \includegraphics[width=\Figsize, height=\Figsize]{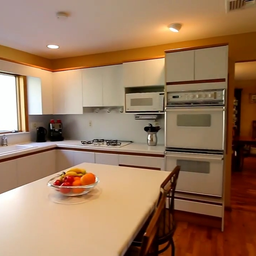} &
        \multicolumn{3}{c}{\betweeninputandoutput \includegraphics[height=\Figsize]{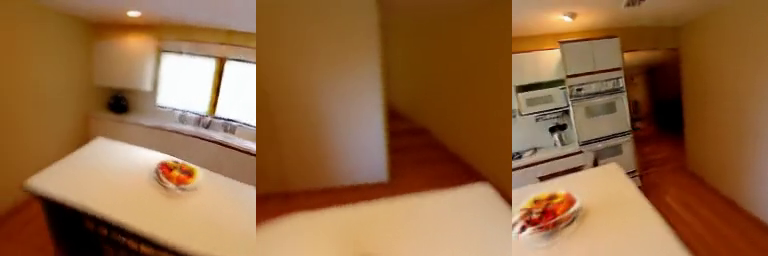}} \\[-1pt]
        
        \beforerow Input view & \multicolumn{3}{c}{\betweeninputandoutput ------------------------ Novel views ------------------------} & 
        Input view & \multicolumn{3}{c}{\betweeninputandoutput ------------------------ Novel views ------------------------} \\[4pt]

        \multicolumn{5}{l}{DTU (Zero-shot)} \\[-1pt]
        \cmidrule(r){1-2}

        \beforerow 
        \includegraphics[width=\Figsize, height=\Figsize]{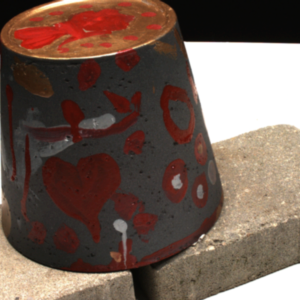} &
        \multicolumn{3}{c}{\betweeninputandoutput \includegraphics[height=\Figsize]{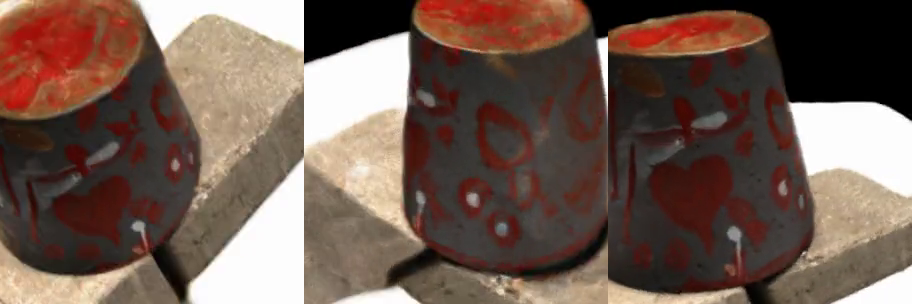}} & 
        
        \betweeninputandoutput \includegraphics[width=\Figsize, height=\Figsize]{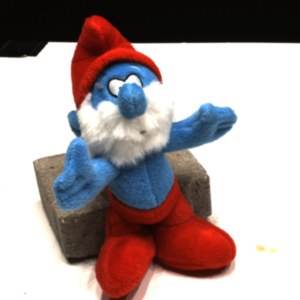} &
        \multicolumn{3}{c}{\betweeninputandoutput \includegraphics[height=\Figsize]{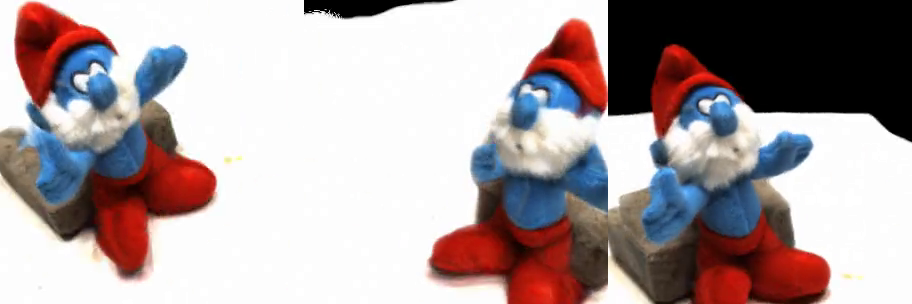}} \\[-1pt]

        \beforerow 
        \includegraphics[width=\Figsize, height=\Figsize]{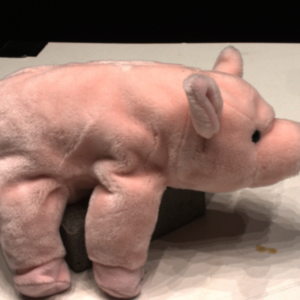} &
        \multicolumn{3}{c}{\betweeninputandoutput \includegraphics[height=\Figsize]{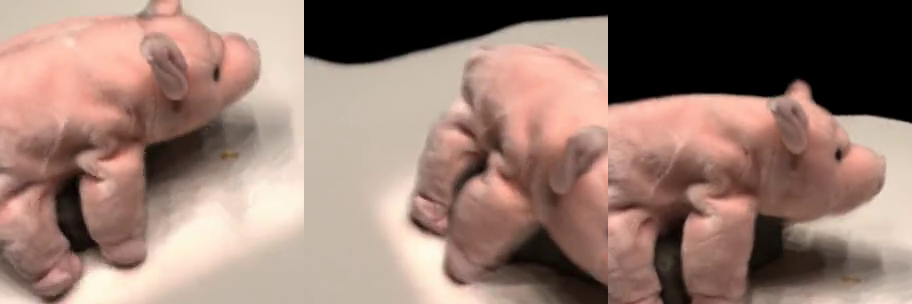}} & 
        
        \betweeninputandoutput \includegraphics[width=\Figsize, height=\Figsize]{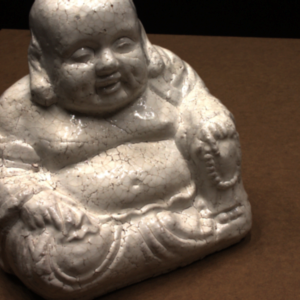} &
        \multicolumn{3}{c}{\betweeninputandoutput \includegraphics[height=\Figsize]{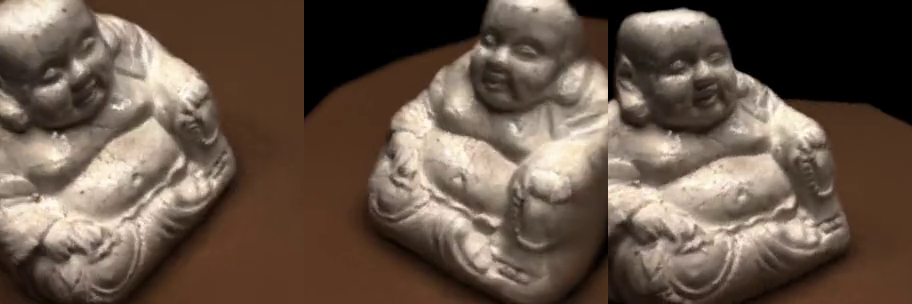}} \\[-1pt]
        
        \beforerow Input view & \multicolumn{3}{c}{\betweeninputandoutput ------------------------ Novel views ------------------------} & 
        Input view & \multicolumn{3}{c}{\betweeninputandoutput ------------------------ Novel views ------------------------} \\[4pt]

        \multicolumn{5}{l}{Mip-NeRF 360 (Zero-shot)} \\[-1pt]
        \cmidrule(r){1-2}

        \beforerow 
        \includegraphics[width=\Figsize, height=\Figsize]{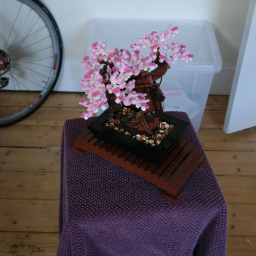} &
        \multicolumn{3}{c}{\betweeninputandoutput \includegraphics[height=\Figsize]{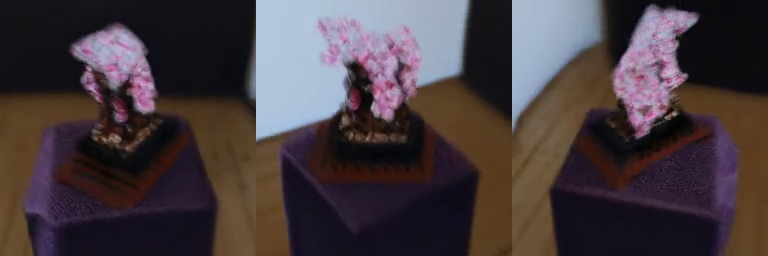}} & 
        
        \betweeninputandoutput \includegraphics[width=\Figsize, height=\Figsize]{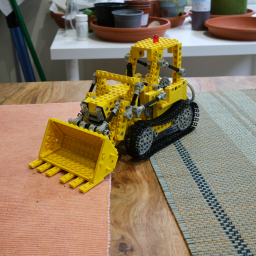} &
        \multicolumn{3}{c}{\betweeninputandoutput \includegraphics[height=\Figsize]{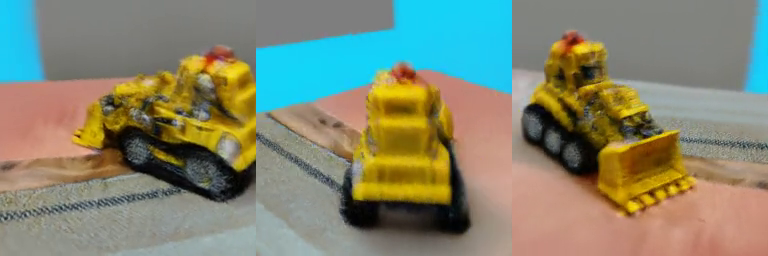}} \\[-1pt]

        \beforerow 
        \includegraphics[width=\Figsize, height=\Figsize]{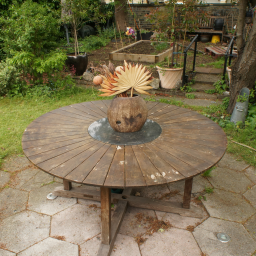} &
        \multicolumn{3}{c}{\betweeninputandoutput \includegraphics[height=\Figsize]{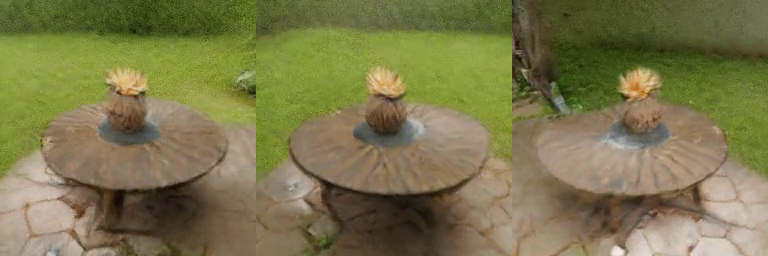}} & 
        
        \betweeninputandoutput \includegraphics[width=\Figsize, height=\Figsize]{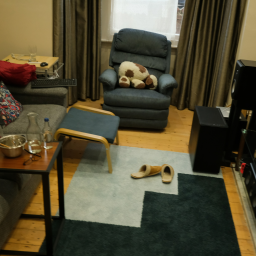} &
        \multicolumn{3}{c}{\betweeninputandoutput \includegraphics[height=\Figsize]{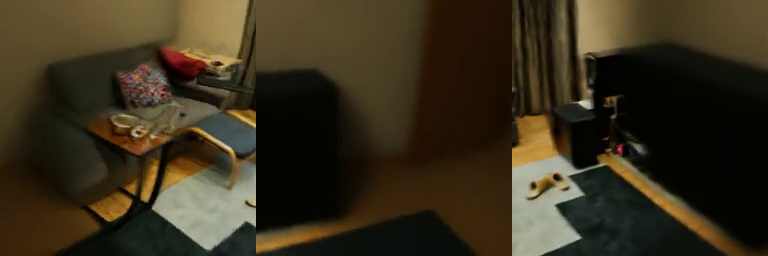}} \\[-1pt]
        
        \beforerow Input view & \multicolumn{3}{c}{\betweeninputandoutput ------------------------ Novel views ------------------------} & 
        Input view & \multicolumn{3}{c}{\betweeninputandoutput ------------------------ Novel views ------------------------} \\[4pt]

        \end{tabular}
        \vspace{-4mm}
        \captionsetup{
          width=1.05\textwidth
        }
        \thispagestyle{empty}
        \caption{Results for view synthesis from a single image. All NeRFs are predicted by the \textit{same} model. 
        }
        \label{fig:result}
\end{figure*}


Models for single-image, 360-degree novel view synthesis (NVS) should produce \emph{realistic} and \emph{diverse} results: the synthesized images should look natural and 3D-consistent to humans, and they should also capture the many possible explanations of unobservable regions. This challenging problem has typically been studied in the context of single objects without backgrounds, where the requirements on both realism and diversity are simplified. Recent progress relies on large 3D datasets like Objaverse-XL \citep{objaversexl} which have enabled training conditional diffusion \citep{zero123} models to perform photorealistic and 3D-consistent NVS via Score Distillation Sampling \citep[SDS;][]{dreamfusion}. Meanwhile, since image diversity mostly lies in the background, not the object, the ignorance of background significantly lowers the expectation of synthesizing diverse images--in fact, most object-centric methods do not consider diversity metrics~\citep{zero123, realfusion, magic123}.

Neither assumption holds for the more challenging problem of zero-shot, 360-degree novel view synthesis on real-world scenes. There is no single, large-scale dataset of scenes with ground-truth geometry, texture, and camera parameters, analogous to Objaverse-XL for objects. 
The background, which cannot be ignored anymore, also needs to be well modeled for synthesizing diverse results. 

We address both issues with our new model, \modelname. Inspired by previous object-centric methods~\citep{zero123, realfusion, magic123}, \modelname~also trains a 2D conditional diffusion model followed by 3D distillation. But unlike them, \modelname~works well on scenes due to two technical innovations: a new camera parametrization and normalization scheme for conditioning, which allows training the diffusion model on diverse scene datasets, and an ``SDS anchoring'' mechanism, improving the background diversity over standard SDS. 

To overcome the key challenge of limited training data, we propose training the diffusion model on a massive mixed dataset comprised of all scenes from CO3D \citep{co3d}, RealEstate10K \citep{realestate10k}, and ACID \citep{acid}, so that the model may potentially handle complex in-the-wild scenes. The mixed data of such scale and diversity are captured with a variety of camera settings and have several different types of 3D ground truth, e.g., computed with COLMAP \citep{colmap} or ORB-SLAM \citep{orbslam}. We show that while the camera conditioning representations from prior methods~\citep{zero123} are too ambiguous or inexpressive to model in-the-wild scenes, our new camera parametrization and normalization scheme allows exploiting such diverse data sources and leads to superior NVS on real-world scenes.
  
Building a 2D conditional diffusion model that works effectively for in-the-wild scenes enables us to then study the limitations of SDS in the scene setting. 
In particular, we observe limited diversity from SDS in the generated scene backgrounds when synthesizing long-range (e.g., 180-degree) novel views. We therefore propose ``SDS anchoring'' to ameliorate the issue. In SDS anchoring, we propose to first sample several ``anchor" novel views using the standard Denoising Diffusion Implicit Model (DDIM) sampling~\citep{song2021denoising}. This yields a collection of pseudo-ground-truth novel views with diverse contents, since DDIM is not prone to mode collapse like SDS. Then, rather than using these views as RGB supervision, we sample from them randomly as conditions for SDS, which enforces diversity while still ensuring 3D-consistent view synthesis. 

\modelname~achieves strong zero-shot generalization to unseen data. We set a new state-of-the-art LPIPS score on the challenging DTU benchmark, even outperforming methods that were directly fine-tuned on this dataset. Since the popular benchmark DTU consists of scenes captured by a forward-facing camera rig and cannot evaluate more challenging pose changes, we propose to use the Mip-NeRF 360 dataset~\citep{mipnerf360} as a single-image novel view synthesis benchmark. \modelname{} achieves the best LPIPS performance on this benchmark. Finally, we show the potential of SDS anchoring for addressing diversity issues in background generation via a user study. 


To summarize, we make the following contributions:
\begin{itemize}
    \item We propose \modelname{}, which enables full-scene NVS from real images. \modelname{} first demonstrates that SDS distillation can be used to lift scenes that are not object-centric and may have complex backgrounds to 3D. 
    
    \item We show that the formulations on handling cameras and scene scale in prior work are either inexpressive or ambiguous for in-the-wild scenes. We propose a new camera conditioning parameterization and a scene normalization scheme. These enable us to train a single model on a large collection of diverse training data consisting of CO3D, RealEstate10K and ACID, allowing strong zero-shot generalization for NVS on in-the-wild images.

    \item We study the limitations of SDS distillation as applied to scenes. Similar to prior work, we identify a diversity issue, which manifests in this case as novel view predictions with monotone backgrounds. We propose SDS anchoring to ameliorate the issue. 
 
    \item We show state-of-the-art LPIPS results on DTU \emph{zero-shot}, surpassing prior methods finetuned on this dataset. Furthermore, we introduce the Mip-NeRF 360 dataset as a scene-level single-image novel view synthesis benchmark and analyze the performances of our and other methods. Finally, we show that our proposed SDS anchoring is preferred via a user study.
\end{itemize}

\section{Related Work}

\noindent\textbf{3D generation.}
DreamFusion \citep{dreamfusion} proposed Score Distillation Sampling (SDS) as a way of leveraging a diffusion model to extract a NeRF given a user-provided text prompt. After DreamFusion, follow-up works such as Magic3D \citep{magic3d}, ATT3D \citep{att3d}, ProlificDreamer \citep{prolificdreamer}, and Fantasia3D \citep{Fantasia3D} improved the quality, diversity, resolution, or run-time. Other types of 3D generative models include GAN-based 3D generative models, which are primarily restricted to single object categories \citep{pigan, giraffe, stylenerf, eg3d, hologan, epigraf} or to synthetic data \citep{get3d}. Recently, 3DGP \citep{3dgp} adapted the GAN approach to train 3D generative models on ImageNet. VQ3D \citep{vq3d} and IVID \citep{ivid} leveraged vector quantization and diffusion, respectively, to learn generative models on ImageNet. One critical critical challenge for scene-based 3D-aware methods 360-degree viewpoint change. Both VQ3D and 3DGP demonstrate only limited camera motion, while IVID generally focuses on small camera motion but can achieve 360-degree views for a small subset of scenes. 


\noindent\textbf{Single-image novel view synthesis.} PixelNeRF \citep{pixelnerf} and DietNeRF \citep{dietnerf} learn to infer NeRFs from sparse views via training an image-based 3D feature extractor or semantic consistency losses, respectively. However, these approaches do not produce renderings resembling crisp natural images from a single image. Several recent diffusion-based approaches achieve high quality results by separating the problem into two stages. First, a (potentially 3D-aware) diffusion model is trained, and second, the diffusion model is used to distill 3D-consistent scene representations given an input image via  techniques like score distillation sampling \citep{dreamfusion}, score Jacobian chaining \citep{sjc}, textual inversion or semantic guidance leveraging the diffusion model \citep{realfusion, nerdi}, or explicit 3D reconstruction from multiple sampled views of the diffusion model \citep{12345, syncdreamer}. Another diffusion-based work, GeNVS \citep{genvs}, achieves 360 camera motion but only for specific object categories such as fire hydrants. By contrast, ZeroNVS generates 360-degree camera motion by default for a variety of scene categories. This is enabled by innovations such as novel camera conditioning representations and SDS anchoring, which enable us to train on massive real scene datasets and then to perform scene-level NVS with up to 360-degree viewpoint change on diverse scene types.

\section{Approach}
\label{sec:approach}
We consider the problem of scene-level novel view synthesis from a single real image. Similar to prior work \citep{zero123, magic123}, we first train a diffusion model $\mathbf{p}_\theta$ to perform novel view synthesis, and then leverage it to perform 3D SDS distillation. Unlike prior work, we focus on scenes rather than objects.
Scenes present several unique challenges. First, prior works use representations for cameras and scale which are either ambiguous or insufficiently expressive for scenes. Second, the inference procedure of prior works is based on SDS, which has a known mode collapse issue and which manifests in scenes through greatly reduced background diversity in predicted views. We will attempt to address these challenges through improved representations and inference procedures for scenes compared with prior work \citep{zero123, magic123}.

We shall begin by introducing some general notation.  Let a scene $S$ consist of a set of images $X = \{X_i\}_{i=1}^n$, depth maps $D=\{D_i\}_{i=1}^n$, extrinsics $E = \{E_i\}_{i=1}^n$, and a shared field-of-view $f$. We note that an extrinsics matrix $E_i$ can be identified with its rotation and translation components, defined by $E_i = (E_i^R, E_i^T)$. We preprocess our data to consist of square images and assume intrinsics are shared within a given scene, and that there is no skew, distortion, or off-center principal point.

\begin{figure}[t]
        \includegraphics[width=.5\textwidth]{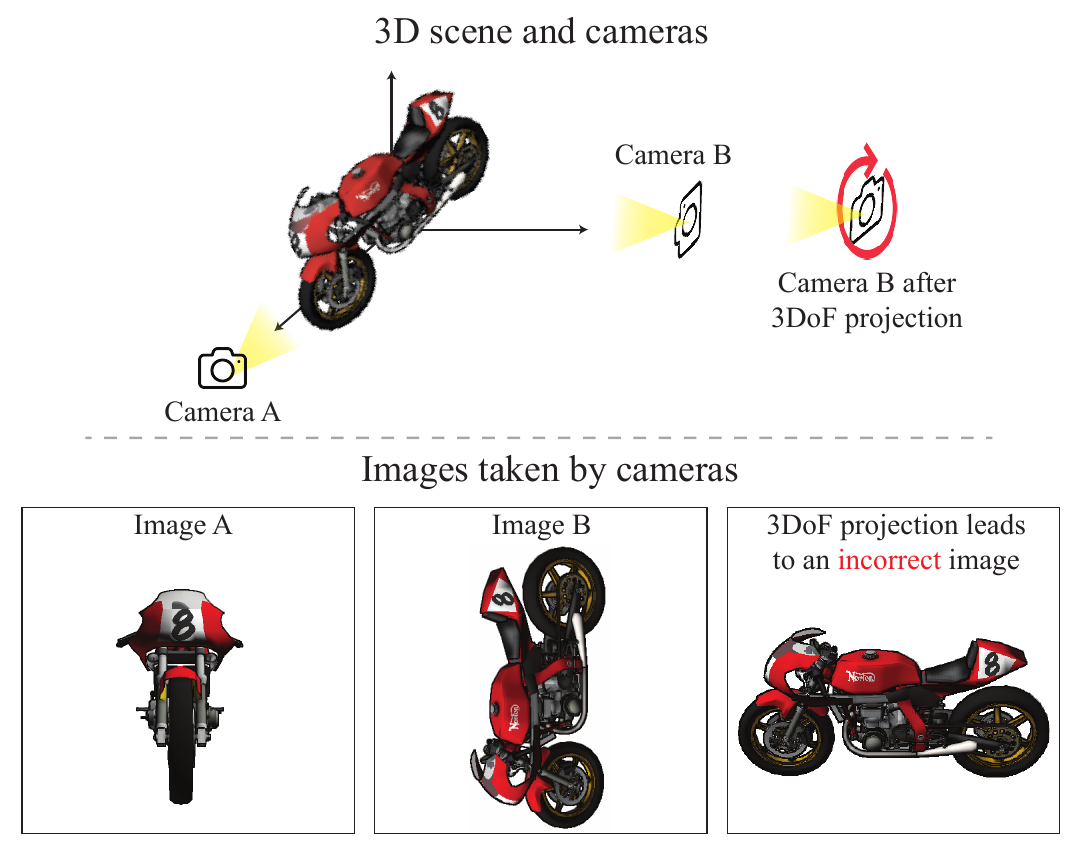}
        \caption{A 3DoF camera pose captures elevation, azimuth, and radius but is incapable of representing a camera's roll (pictured) or cameras positioned and oriented arbitrarily in space. 
        }
        \label{fig:3dof_failure}
\end{figure}

\begin{figure}[t]
\centering
    \includegraphics[width=.4\textwidth]{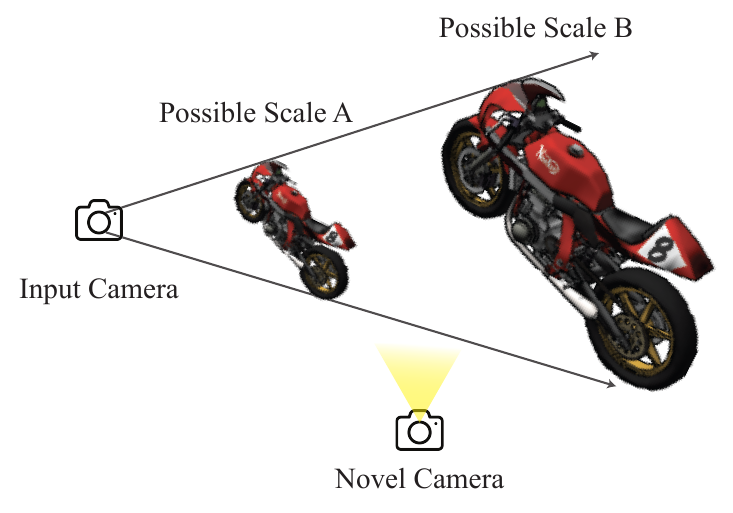}
    \caption{
    To a monocular camera, a small object close to the camera (left) and a large object at a distance (right) appear identical, despite representing different scenes. Scale ambiguity in input view leads to multiple plausible novel views. 
    }
    \label{fig:camera_explanation_1}
    \vspace{-5mm}
\end{figure}

\begin{figure}
        \includegraphics[width=.49\textwidth]{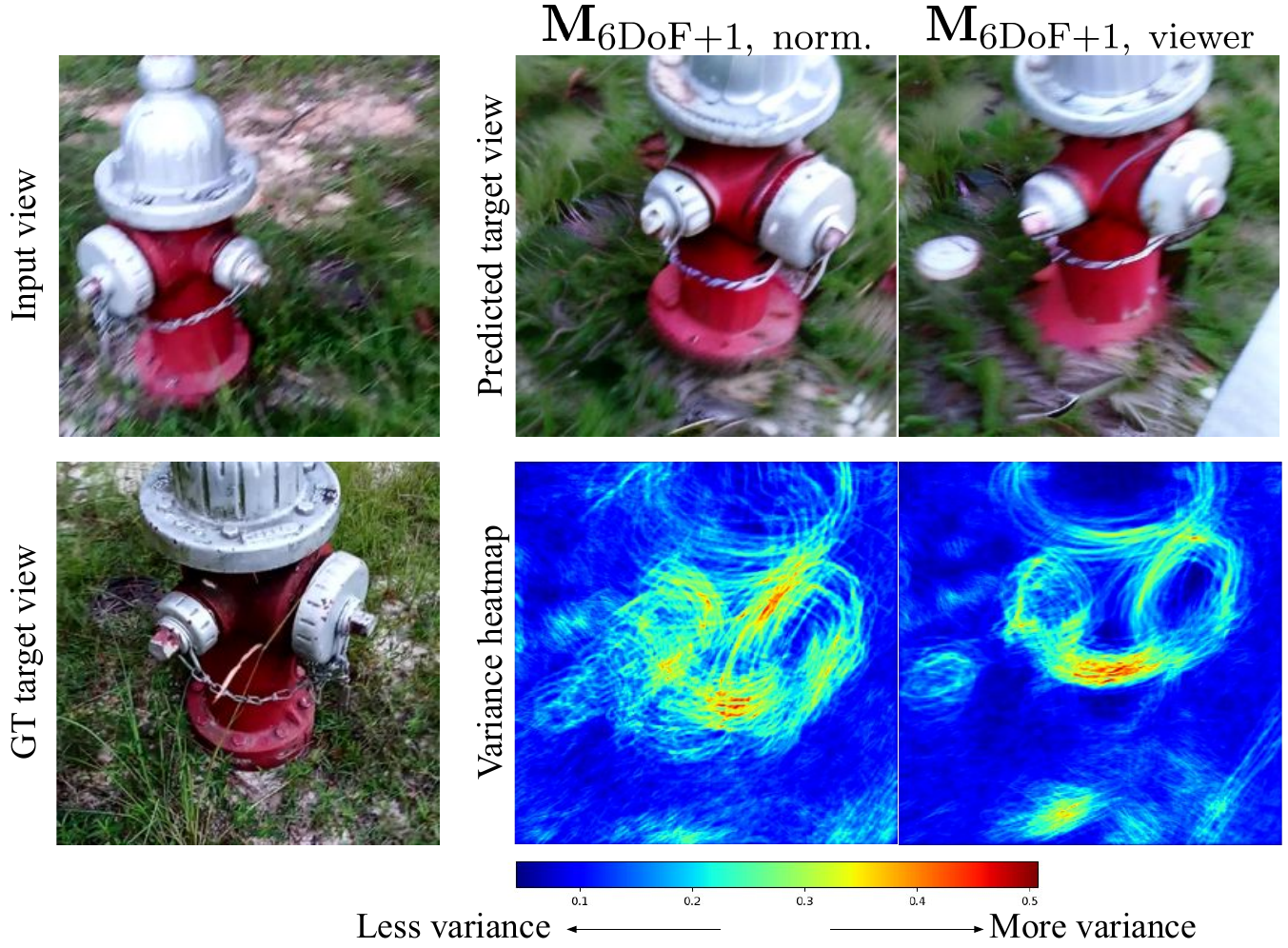}
        \caption{Samples and variance heatmaps of the Sobel edges of multiple samples from ZeroNVS. $\mathbf{M}_{\mathrm{6DoF+1,~viewer}}$ reduces randomness from scale ambiguity.}
        \label{fig:normalization_explanation}
    \vspace{-6mm}
\end{figure}

We will focus on the design of the conditional information which is passed to the view synthesis diffusion model $\mathbf{p}_\theta$ in addition to the input image. This conditional information can be represented via a function, $\mathbf{M}(D,f,E,i,j)$, which computes a conditioning embedding given the depth maps and extrinsics for the scene, the field of view, and the indices $i,j$ of the input and target view respectively. We learn a generative model over novel views $\mathbf{p_\theta}$ given by
\[ 
    X_{j} \sim \mathbf{p_\theta}(X_{j} | X_{i}, \mathbf{M}(D,f,E,i,j))~.
\]
The output of $\mathbf{M}$ and the input image $X_{i}$ are the only information used by the model for NVS. Both Zero-1-to-3 (Section \ref{sec:representing_zero123}) and our model, as well as several intermediate models that we will study (Sections \ref{sec:representing_generic} and \ref{sec:method_normalization}), can be regarded as different choices for $\mathbf{M}$. As we illustrate in Figures \ref{fig:3dof_failure}, \ref{fig:camera_explanation_1}, \ref{fig:normalization_explanation} and 
\ref{fig:camera_explanation_2}, and verify in experiments, different choices for $\mathbf{M}$ can have drastic impacts on performance. 

\begin{figure*}[t]
\vspace{-12pt}
\includegraphics[width=\textwidth]{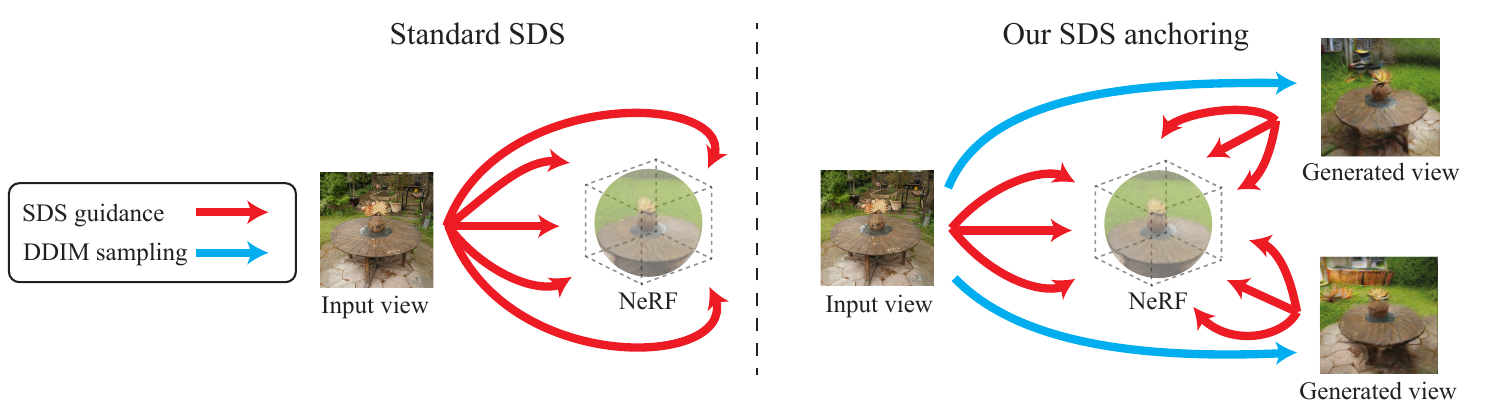}
\vspace{-6mm}
\caption{SDS-based NeRF distillation (left) uses the same guidance image for all 360 degrees of novel views. Our ``SDS anchoring'' (right) first samples novel views via DDIM \citep{ddim}, and then uses the nearest image (whether the input or a sampled novel view) for guidance.}
\label{fig:sds_anchoring}
\vspace{-6mm}
\end{figure*}



\subsection{Representing Objects for View Synthesis}
\label{sec:representing_zero123}
Zero-1-to-3 \citep{zero123} represents poses with 3 degrees of freedom, given by an elevation $\theta$, azimuth $\phi$, and radius $z$. Let $\mathbf{P}:\mathrm{SE}(3) \rightarrow \mathbb{R}^3$ be the projection to $(\theta,\phi,z)$, then
\[
\mathbf{M}_{\mathrm{Zero-1-to-3}}(D,f,E,i,j) = \mathbf{P}(E_{i}) - \mathbf{P}(E_{j})
\]
is the camera conditioning representation used by Zero-1-to-3. For object mesh datasets such as Objaverse \citep{objaverse} and Objaverse-XL \citep{objaversexl}, this representation is appropriate because the data is known to consist of single objects without backgrounds, aligned and centered at the origin and imaged from training cameras generated with three degrees of freedom. However, such a parameterization limits the model's ability to generalize to non-object-centric images, and to real-world data.
In real-world data, poses can have six degrees of freedom, incorporating both rotation (pitch, roll, yaw) and 3D translation. An illustration of a failure of the 3DoF camera representation is shown in Figure \ref{fig:3dof_failure}. 

\subsection{Representing Scenes for View Synthesis} \label{sec:representing_generic}

For scenes, we should use a camera representation with six degrees of freedom that can capture all possible positions and orientations. One straightforward choice is the relative pose parameterization \citep{3dim}. We propose to also include the field of view as an additional degree of freedom. We term this combined representation ``6DoF+1''. This gives us
\[
\mathbf{M}_{\mathrm{6DoF+1}}(D,f,E,i, j) = [E_{i}^{-1}E_{j},f].
\]

Importantly, $\mathbf{M}_{\mathrm{6DoF+1}}$ is invariant to any rigid transformation $\tilde E$ of the scene, so that we have 
\[
\mathbf{M}_{\mathrm{6DoF+1}}(D,f,\tilde E\cdot E,i, j) = [E_{i}^{-1}E_{j},f]~.
\]
This is useful given the arbitrary nature of the poses for our datasets which are determined by COLMAP or ORB-SLAM. The poses discovered via these algorithms are not related to any meaningful alignment of the scene, such as a rigid transformation and scale transformation which align the scene to some canonical frame and unit of scale. Although we have seen that $\mathbf{M}_{\mathrm{6DoF+1}}$ is invariant to rigid transformations of the scene, it is not invariant to scale. The scene scales determined by COLMAP and ORB-SLAM are also arbitrary and may vary significantly. One solution is to directly normalize the camera locations. Let $\mathbf{R}(E, \lambda): \textrm{SE}(3) \times \mathbb{R} \rightarrow \textrm{SE}(3)$ be a function that scales the translation component of $E$ by $\lambda$. Then we define
\begin{align*}
s =& \frac{1}{n} \sum\limits_{i=1}^n \|E_i^T - \frac{1}{n} \sum\limits_{j=1}^n E_j^T\|_2 ~,\\
\mathbf{M}_{\mathrm{6DoF+1,~norm.}}(D,f,E,i, j) 
=& \Big[
  \mathbf{R}\Big(E_{i}^{-1}E_j, \frac{1}{s}\Big), f
  \Big] ~,
\end{align*}
where $s$ is the average norm of the camera locations when the mean of the camera locations is chosen as the origin. In $\mathbf{M}_{\mathrm{6DoF+1,~norm.}}$, the camera locations are normalized via rescaling by $\frac{1}{s}$, in contrast to $\mathbf{M}_{\mathrm{6DoF+1}}$ where the scales are arbitrary. This choice of $\mathbf{M}$ assures that scenes from our mixture of datasets will have similar scales. 

\vspace{-2mm}
\subsection{Addressing Scale Ambiguity with a New \\Normalization Scheme}
\label{sec:method_normalization}

\begin{figure}
        \centering
        \includegraphics[width=.49\textwidth]{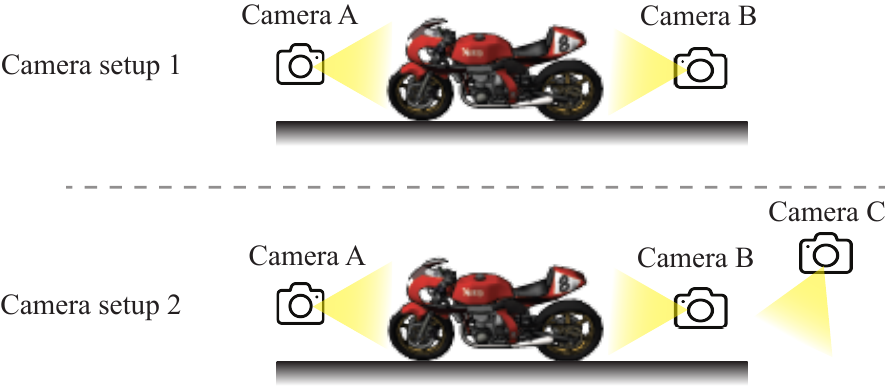}
        \caption{Top: A scene with two cameras facing the object. Bottom: The same scene with a new camera added facing the ground. Addition of Camera C under $\mathbf{M}_{\mathrm{6DoF+1,~agg.}}$ drastically changes the scene’s scale. $\mathbf{M}_{\mathrm{6DoF+1,~viewer.}}$ avoids this.
        }
        \label{fig:camera_explanation_2}
    \vspace{-5mm}
\end{figure}

The representation $\mathbf{M}_{\mathrm{6DoF+1,~norm.}}$ achieves reasonable performance on real scenes by addressing issues in prior representations with limited degrees of freedom and handling of scale. However, a normalization scheme that better addresses scale ambiguity may lead to improved performance. Scene scale is ambiguous given a monocular input image \citep{ranftl2022midas, leres}. This complicates NVS, as we illustrate in Figure~\ref{fig:camera_explanation_1}.
We therefore choose to introduce information about the scale of the visible content to our conditioning embedding function $\mathbf{M}$. 
Rather than normalize by camera locations, Stereo Magnification \citep{realestate10k} takes the 5-th quantile of each depth map of the scene, and then takes the 10-th quantile of this aggregated set of numbers, and declares this as the scene scale. Let $\mathbf{Q}_k$
be a function which takes the $k$-th quantile of a set of numbers, then we define
\begin{align*}
q =& \mathbf{Q}_{10}(\{\mathbf{Q}_{5}(D_i)\}_{i=1}^n) ~,\\
\mathbf{M}_{\mathrm{6DoF+1,~agg.}}(D,f,E,i, j) =& 
\Big[\mathbf{R}\Big(
    E_{i}^{-1}E_j,
    \frac{1}{q}
    \Big),f\Big]~,
\end{align*}
where in $\mathbf{M}_{\mathrm{6DoF+1,~agg.}}$, $q$ is the scale applied to the translation component of the scene's cameras before computing the relative pose. In this way $\mathbf{M}_{\mathrm{6DoF+1,~agg.}}$ is different from $\mathbf{M}_{\mathrm{6DoF+1,~norm.}}$ because the camera conditioning representation contains information about the scale of the visible content from the depth maps $D_i$. Although conditioning with $\mathbf{M}_{\mathrm{6DoF+1,~agg.}}$ improves performance, there are two issues. The first arises from aggregating the quantiles over all the images. In Figure \ref{fig:camera_explanation_2}, adding an additional Camera C to the scene changes the value of $\mathbf{M}_{\mathrm{6DoF+1,~agg.}}$ despite nothing else having changed about the scene. This makes the view synthesis task from either Camera A or Camera B more ambiguous. To ensure this is impossible, we can simply eliminate the aggregation step over the quantiles of all depth maps in the scene. The second issue arises from different depth statistics within the mixture of datasets we use for training. ORB-SLAM generally produces sparser depth maps than COLMAP, and therefore the value of $\mathbf{Q}_k$ may have different meanings for each. We therefore use an off-the-shelf depth estimator \citep{ranftl2021dpt} to fill holes in the depth maps. We denote the depth $D_i$ infilled in this way as $\bar D_i$. We then apply $\mathbf{Q}_k$ to dense depth maps $\bar D_i$ instead. We emphasize that the depth estimator is \emph{not} used during inference or distillation. Its purpose is only for the model to learn a consistent definition of scale during training. These two fixes lead to our proposed normalization, which is fully viewer-centric. We define it as
\begin{align*}
q_{i} =& \mathbf{Q}_{20}(\bar D_{i}) ~,\\
\mathbf{M}_{\mathrm{6DoF+1,~viewer}}(D,f,E,i, j) =& 
\Big[\mathbf{R}\Big(
    E_{i}^{-1}E_{j},
    \frac{1}{q_{i}}
    \Big),f\Big]~,
\end{align*}
where in $\mathbf{M}_{\mathrm{6DoF+1,~viewer}}$, the scale $q_i$ applied to the cameras is dependent only on the depth map in the input view $\bar D_i$, different from $\mathbf{M}_{\mathrm{6DoF+1,~agg.}}$ where the scale $q$ computed by aggregating over all $D_i$. At inference the value of $q_{i}$ can be chosen heuristically without compromising performance.  Correcting for the scale ambiguities in this way improves metrics, which we show in Section~\ref{sec:experiments}. 

\begin{figure*}
     \begin{centering}\small
     \setlength{\tabcolsep}{1pt}
     \newcommand{\figwidth}{.24\textwidth}
\begin{tabular}{cccc}
     \includegraphics[width=\figwidth]{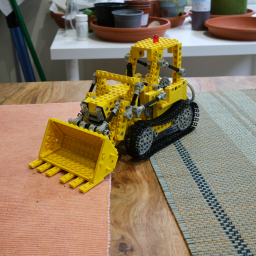} &
     
     \includegraphics[width=\figwidth]{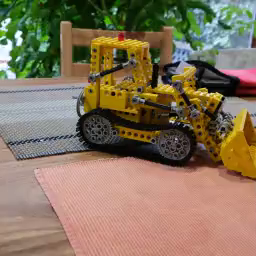} &
     
     \includegraphics[width=\figwidth]{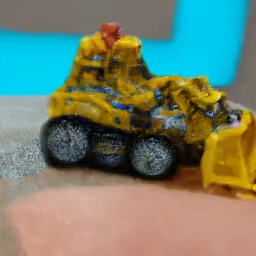} &
     \includegraphics[width=\figwidth]{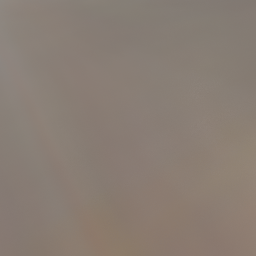} \\
     
     Input view & GT novel view & \modelname~(ours) &  PixelNeRF \\
      &&  PSNR=10.8, SSIM=0.22 &  PSNR=12.2, SSIM=0.30  \\
\end{tabular}
\vspace{-3mm}
\caption{Limitations of PSNR and SSIM for view synthesis evaluation. Misalignments can lead to worse PSNR and SSIM values for predictions that are more semantically sensible. 
}
\label{fig:psnr_limitations}
\end{centering}
\end{figure*}

\begin{table}
    \small
    \centering
        \begin{tabular}{lccc}
        \toprule
        \textbf{NVS on DTU} & \textbf{LPIPS} $\downarrow$ & \textbf{PSNR} $\uparrow$ & \textbf{SSIM} $\uparrow$ \\
        \midrule
        $\textrm{DS-NeRF}^\dagger$ & 0.649 & 12.17 & 0.410\\
        PixelNeRF & 0.535 & 15.55 & 0.537   \\
        SinNeRF & 0.525 & \textbf{16.52} & \textbf{0.560}  \\
        DietNeRF & 0.487 & 14.24 & 0.481   \\
        NeRDi & 0.421 & 14.47 & 0.465   \\
        \midrule
        \modelname~(ours) & 
        \textbf{0.380} & 13.55 & 0.469 \\
        \bottomrule
        \end{tabular}
        \vspace{-3mm}
        \caption{\textbf{Comparison with the state of the art.} We set a new state-of-the-art for LPIPS on DTU despite being the only method not fine-tuned on DTU. $\dagger=$Performance reported in \citet{sinnerf}.
        }
        \label{tab:dtu_nvs}
\vspace{-4mm}
\end{table}

\begin{table}
\small
\centering
        \begin{tabular}{lcccc}
        \toprule
        \textbf{NVS} & \textbf{LPIPS} $\downarrow$ & \textbf{PSNR} $\uparrow$ & \textbf{SSIM} $\uparrow$ \\
        \midrule
        \textbf{Mip-NeRF 360 Dataset} & \\
        Zero-1-to-3 & 0.667 & 11.7 & 0.196  \\
        PixelNeRF & 0.718 & \textbf{16.5} & \textbf{0.556} \\
        \modelname~(ours) & \textbf{0.625} & 13.2 & 0.240  \\
        \midrule
        \textbf{DTU Dataset} & \\
        Zero-1-to-3 & 0.472 & 10.70 & 0.383 \\
        PixelNeRF & 0.738 & 10.46 & 0.397  \\
        \modelname~(ours) & 
        \textbf{0.380} & \textbf{13.55} & \textbf{0.469} \\
        \bottomrule
        \end{tabular}
        \vspace{-3mm}
        \caption{\textbf{Zero-shot comparison}. Comparison with baselines re-trained on our mixture dataset. ZeroNVS outperforms Zero-1-to-3 even when Zero-1-to-3 is trained on the same scene data. Extensive video comparisons are in the supplementary. 
        }
        \label{tab:zero_shot}
        \vspace{-4mm}
\end{table}

\subsection{Improving Diversity with SDS Anchoring}

Diffusion models trained with the improved camera conditioning representation $\mathbf{M}_{\mathrm{6DoF+1,~viewer}}$ achieve superior view synthesis results via 3D SDS distillation. However, for large viewpoint changes, novel view synthesis is also a generation problem, and it may be desirable to generate diverse and plausible contents rather than contents that are only optimal on average for metrics such as PSNR, SSIM, and LPIPS. However, \citet{dreamfusion} noted that even when the underlying generative model produces diverse images, SDS distillation of that model tends to seek a single mode. For novel view synthesis of scenes via SDS, we observe a unique manifestation of this diversity issue: lack of diversity is especially apparent in inferred backgrounds. Often, SDS distillation predicts a gray or monotone background for regions not observed by the input camera. 

\begin{figure*}
     \setlength{\tabcolsep}{2pt}
\begin{tabular}{cccc}
     \includegraphics[width=\textwidth]{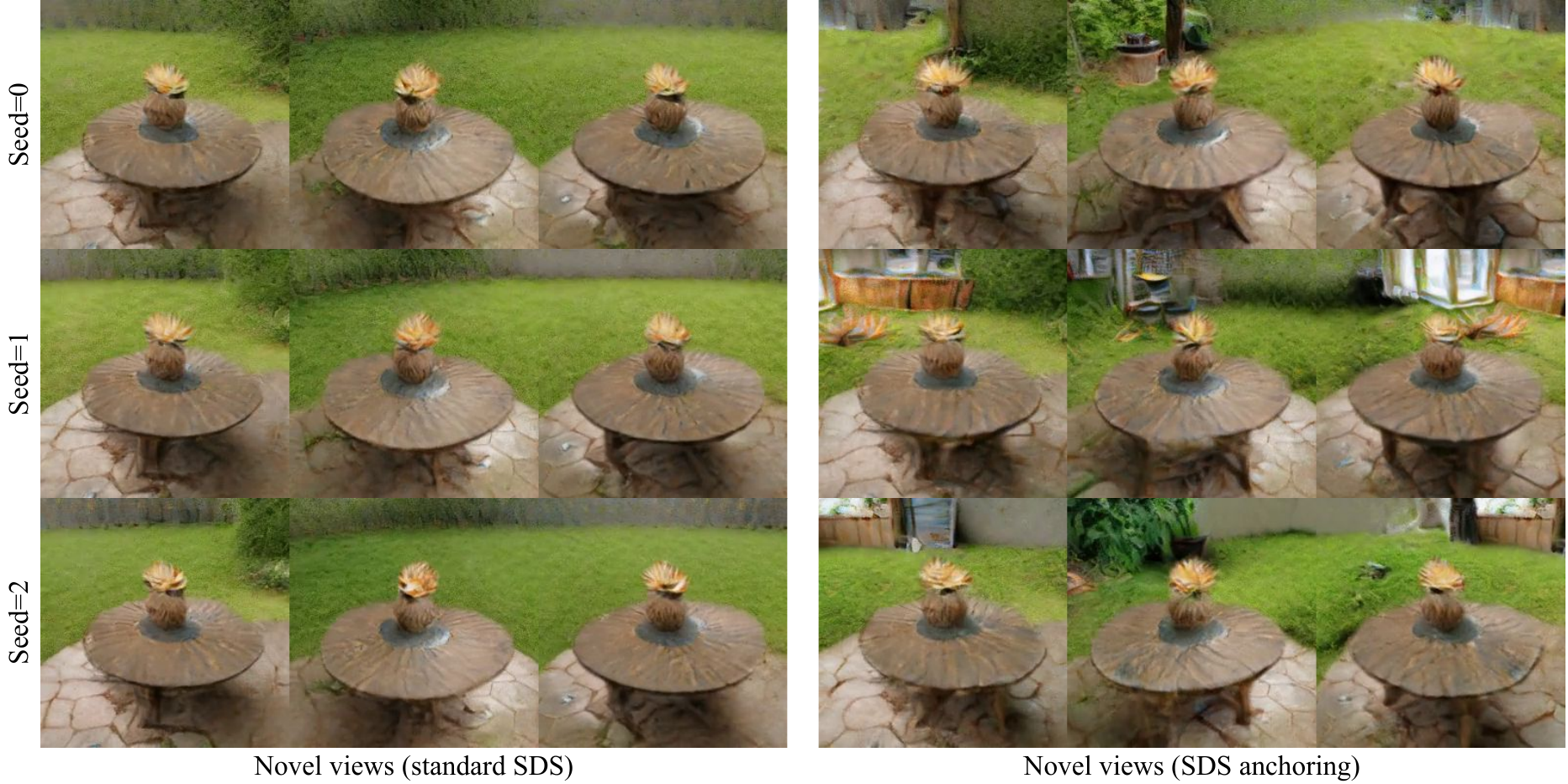} & 
\end{tabular}
\vspace{-4mm}
\caption{Whereas standard SDS (left) tends to predict monotonous backgrounds, our SDS anchoring (right) generates more diverse background contents. Additionally, SDS anchoring generates noticeably different results depending on the seed.
}
\vspace{-2mm}
\label{fig:anchoring_qualitative}
\end{figure*}

To remedy this, we propose ``SDS anchoring'' (Figure \ref{fig:sds_anchoring}). 
With SDS anchoring, we first directly sample $k$ novel views $\boldsymbol{\hat X}_{k} = \{\hat X_j\}_{j=1}^k$ with $\hat X_j \sim p(X_j | X_{i}, \mathbf{M}(D, f, E, i, j))$ from poses evenly spaced in azimuth for maximum scene coverage. We sample the novel views via DDIM \citep{ddim}, which does not have the mode collapse issues of SDS. Each novel view is generated conditional on the input view. Then, when optimizing the SDS objective, we condition the diffusion model not on the input view, but on the nearest view.
As shown quantitatively in a user study in Section \ref{sec:experiments} and qualitatively in Figure \ref{fig:anchoring_qualitative}, SDS anchoring produces more diverse background contents. We provide more details about the setup of SDS anchoring in the supplementary.

\section{Experiments}
\label{sec:experiments}

\begin{figure*}
     \setlength{\tabcolsep}{2pt}
     \small
     \newcommand{\figwidth}{.24\textwidth}
\begin{tabular}{cccc}
     \includegraphics[width=\figwidth]{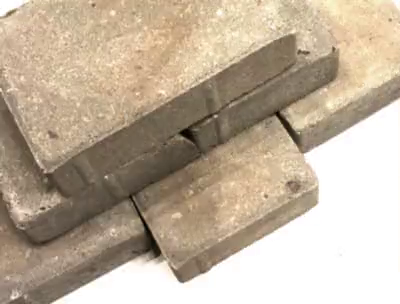} &
     
     \includegraphics[width=\figwidth]{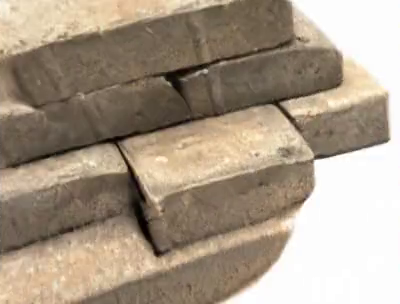} &
     
     \includegraphics[width=\figwidth]{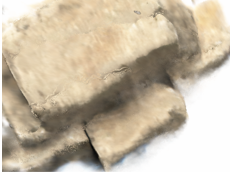} &
     \includegraphics[width=\figwidth]{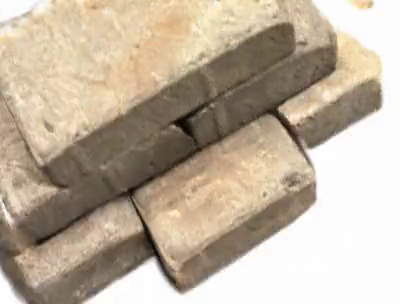} \\

     \includegraphics[width=\figwidth]{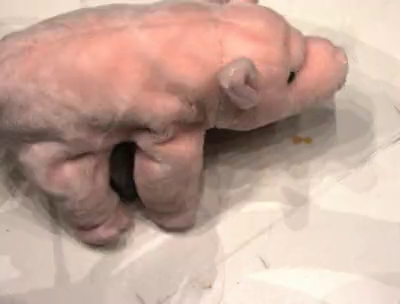} &
     
     \includegraphics[width=\figwidth]{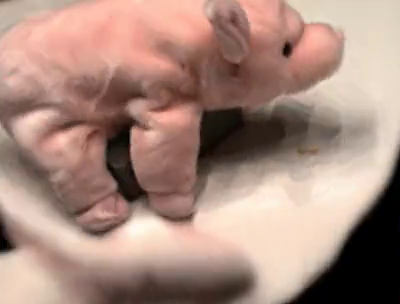} &
     
     \includegraphics[width=\figwidth]{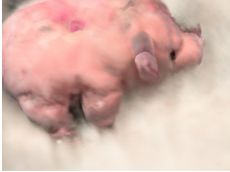} &
     \includegraphics[width=\figwidth]{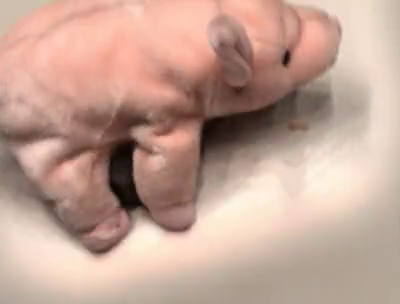} \\
    GT novel view & Zero-1-to-3 & NerDi & \modelname~(ours) \\
\end{tabular}
\vspace{-3mm}
\caption{Qualitative comparison between baseline methods and our method.}
\vspace{-5mm}
\label{fig:qualitative_comparison}
\end{figure*}

\begin{table}
    \centering
    \begin{tabular}{lccc}
    \toprule
    \textbf{NVS on DTU} & \textbf{LPIPS} $\downarrow$ & \textbf{PSNR} $\uparrow$ & \textbf{SSIM} $\uparrow$ \\
    \midrule
    All datasets & \textbf{0.421} & \textbf{12.2} & \textbf{0.444} \\
    -ACID & 0.446 & 11.5 & 0.405 \\
    -CO3D & 0.456 & 10.7 & 0.407 \\
    -RealEstate10K & 0.435 & 12.0 & 0.429 \\
    \bottomrule
    \end{tabular}
    \vspace{-3mm}
    \caption{\textbf{Ablation study on training data.} Training on all datasets improves performance.}
    \label{tab:dataset_ablation}
    \vspace{-6mm}
\end{table}

\begin{table*}
    \setlength\tabcolsep{4.0pt}
    \renewcommand\arraystretch{1.2}
    \centering
    \small
    \begin{tabular}{lrrrrrrrrrrrr}
        \toprule
        & \multicolumn{9}{c}{\textbf{2D novel view synthesis}} & \multicolumn{3}{c}{\textbf{3D NeRF distillation}}\\
        \cmidrule(lr){2-10} \cmidrule(lr){11-13}
        & \multicolumn{3}{c}{\textbf{CO3D}} & \multicolumn{3}{c}{\textbf{RealEstate10K}} & \multicolumn{3}{c}{\textbf{ACID}} & \multicolumn{3}{c}{\textbf{DTU}} \\
        \cmidrule(lr){2-4}\cmidrule(lr){5-7}\cmidrule(lr){8-10}\cmidrule(lr){11-13}
        \textbf{Conditioning} & \textbf{PSNR} & \textbf{SSIM} & \textbf{LPIPS}& \textbf{PSNR} & \textbf{SSIM} & \textbf{LPIPS}& \textbf{PSNR} & \textbf{SSIM} & \textbf{LPIPS}& \textbf{PSNR} & \textbf{SSIM} & \textbf{LPIPS} \\
        \midrule
        $\mathbf{M}_{\mathrm{Zero-1-to-3}}$
        & 12.0 & .366 & .590
        & 11.7 & .338 & .534
        & 15.5 & .371 & .431 

        & 10.3 & .384 & .477 \\
        
        $\mathbf{M}_{\mathrm{6DoF+1}}$ 
        & 12.2 & .370 & .575 
        & 12.5 & .380 & .483 
        & 15.2 & .363 & .445

        & 9.5 & .347 & .472 \\
        
        $\mathbf{M}_{\mathrm{6DoF+1,~norm.}}$    
        & 12.9 & .392 & .542
        & 12.9 & .408 & .450
        & 16.5 & .398 & .398

        & 11.5 & .422 & .421 \\
        
        $\mathbf{M}_{\mathrm{6DoF+1,~agg.}}$
        & 13.2 & .402 & .527
        & \textbf{13.5} & \textbf{.441} & .417
        & 16.9 & .411 & .378 

        & \textbf{12.2} & .436 & \textbf{.420} \\
        
        $\mathbf{M}_{\mathrm{6DoF+1,~viewer}}$
        & \textbf{13.4} & \textbf{.407} & \textbf{.515} 
        & \textbf{13.5} & .440 & \textbf{.414} 
        & \textbf{17.1} & \textbf{.415} & \textbf{.368} 

        & \textbf{12.2} & \textbf{.444} & .421 \\
        \bottomrule
    \end{tabular}
    \vspace{-6pt}
    \caption{\textbf{Ablation study on the conditioning representation $\mathbf{M}$.} Our $\mathbf{M}_{\mathrm{6DoF+1,~viewer}}$ matches or outperforms other representations.}
    \label{tab:conditioning_ablation}
    \vspace{-5mm}
\end{table*}

\subsection{Setup}
\noindent\textbf{Datasets.} Our models are trained on a mixture dataset consisting of CO3D \citep{co3d}, ACID \citep{acid}, and RealEstate10K \citep{realestate10k}. Each example is sampled uniformly at random from the three datasets. We train at $256 \times 256$ resolution, center-cropping and adjusting the intrinsics for each image and scene as necessary. We train using our representation $\mathbf{M}_{\mathrm{6DoF+1,~viewer}}$ unless otherwise specified. 
We provide more training details in the supplementary.

We evaluate our trained diffusion models on held-out subsets of CO3D, ACID, and RealEstate10K respectively, for 2D novel view synthesis. Our main evaluations are for zero-shot 3D consistent novel view synthesis, where we compare against other techniques on the DTU benchmark \citep{dtu} and on the Mip-NeRF 360 dataset \citep{mipnerf360}. We evaluate at $256 \times 256$ resolution except for DTU, for which we use $400 \times 300$ resolution to be comparable to prior art. 

\noindent\textbf{Implementation details.} Our diffusion model training code is written in PyTorch and based on the public code for Zero-1-to-3 \citep{zero123}. We initialize from the pretrained Zero-1-to-3-XL, swapping out the conditioning module to accommodate our novel parameterizations. Our distillation code is implemented in Threestudio \citep{threestudio}. We use a custom NeRF network combining features of Mip-NeRF 360 with Instant-NGP \citep{instantngp}. The noise schedule is annealed following \citet{prolificdreamer}. For details please see the supplementary.

\subsection{Main Results}

We evaluate all methods using the standard set of novel view synthesis metrics: PSNR, SSIM, and LPIPS. 
We weigh LPIPS more heavily in the comparison due to the well-known issues with PSNR and SSIM as discussed in~\citep{nerdi,genvs}. 
We confirm that PSNR and SSIM do not correlate well with performance in our setting, as illustrated in Figure~\ref{fig:psnr_limitations}.

The results are shown in Table~\ref{tab:dtu_nvs}. 
We first compare against baseline methods DS-NeRF \citep{dsnerf}, PixelNeRF \citep{pixelnerf}, SinNeRF \citep{sinnerf}, DietNeRF \citep{dietnerf}, and NeRDi \citep{nerdi} on DTU. All these methods are trained on DTU, but we achieve a state-of-the-art LPIPS despite being fully zero-shot. We show visual comparisons in Figure \ref{fig:qualitative_comparison}.

DTU scenes are limited to relatively simple forward-facing scenes. Therefore, we introduce a more challenging benchmark dataset, the Mip-NeRF 360 dataset, to benchmark the task of 360-degree view synthesis from a single image. We use this benchmark as a zero-shot benchmark, and train three baseline models on our mixture dataset to compare zero-shot performance. Our method is the best on LPIPS for this dataset. On DTU, we exceed Zero-1-to-3 and the zero-shot PixelNeRF model on all metrics, not just LPIPS. Performance is shown in Table \ref{tab:zero_shot}. All numbers for our method and Zero-1-to-3 are for NeRFs predicted from SDS distillation unless otherwise noted.

Limited diversity is a known issue with SDS-based methods, but the long run time of SDS-based methods makes typical generation-based metrics such as FID cost-prohibitive. Therefore, we quantify the improved diversity from SDS anchoring via a user study of 21 users on the Mip-NeRF 360 dataset. Users were asked to compare scenes predicted with and without SDS anchoring along three dimensions: Realism, Creativity, and Overall Preference. The preferences for SDS anchoring were: Realism (78\%), Creativity (82\%), and Overall Preference (80\%). The supplementary provides more details about the setup of the study. Figure~\ref{fig:anchoring_qualitative} includes qualitative examples that show the advantages of SDS anchoring, and the supplementary webpage contains the videos which were shown in the study. 

    



We conduct multiple ablations to verify our contributions. We verify the benefits of each of our multiple multiview scene datasets in Table \ref{tab:dataset_ablation}. Removing any of the three datasets on which \modelname{} is trained reduces performance. In Table \ref{tab:conditioning_ablation}, we analyze the diffusion model's performance on held-out subsets of our datasets, with the various parameterizations discussed in Section~\ref{sec:approach}. We see that as the conditioning parameterization is further refined, the performance continues to increase. Due to computational constraints, we train the ablation diffusion models for fewer steps than our main model, hence the slightly worse performance relative to Table \ref{tab:dtu_nvs}. We provide more details in the supplementary.

\vspace{-1mm}
\section{Conclusion} 

We have introduced \modelname, a system for 3D-consistent novel view synthesis from a single image for generic scenes. We showed its state-of-the-art performance on existing NVS benchmarks and proposed the Mip-NeRF 360 dataset as a more challenging benchmark for single-image NVS.


\vspace{-10pt}
\paragraph{Acknowledgments.}
The work is in part supported by NSF CCRI \#2120095, RI \#2211258, ONR MURI N00014-22-1-2740, and Google.

{
    \small
    \bibliographystyle{ieeenat_fullname}
    \bibliography{main}
}

\clearpage

\appendix




\section{Details: Diffusion model training}
\label{appendix:training}

\subsection{Model}

We train diffusion models for various camera conditioning parameterizations:
$\mathbf{M}_{\mathrm{Zero-1-to-3}}$,
$\mathbf{M}_{\mathrm{6DoF+1}}$,
$\mathbf{M}_{\mathrm{6DoF+1,~norm.}}$,
$\mathbf{M}_{\mathrm{6DoF+1,~agg.}}$, and
$\mathbf{M}_{\mathrm{6DoF+1,~viewer}}$. Our runtime is identical to Zero-1-to-3 \citep{zero123} as the camera conditioning novelties we introduce add negligible overhead and can be done mainly in the dataloader. We train our main model for $60,000$ steps with batch size $1536$. We find that performance tends to saturate after about $20,000$ steps for all models, though it does not decrease. For inference of the 2D diffusion model, we use $500$ DDIM steps and guidance scale $3.0$. 

\noindent\paragraph{Details for}\hspace{-2mm}$\mathbf{M}_{\mathrm{6DoF+1}}$: To embed the field of view $f$ in radians, we use a $3$-dimensional vector consisting of $[f, \sin(f), \cos(f)].$ When concatenated with the $4\times 4=16$-dimensional relative pose matrix, this gives a $19$-dimensional conditioning vector.

\noindent\paragraph{Details for}\hspace{-2mm}$\mathbf{M}_{\mathrm{6DoF+1,~viewer}}$: We use the DPT-SwinV2-256 depth model \citep{ranftl2021dpt} to infill depth maps from ORB-SLAM and COLMAP on the ACID, RealEstate10K, and CO3D datasets. We infill the invalid depth map regions only after aligning the disparity from the monodepth estimator to the ground-truth sparse depth map via the optimal scale and shift following \citet{ranftl2022midas}. We downsample the depth map $4\times$ so that the quantile function is evaluated quickly. 

At inference time, the value of $\mathbf{Q}_{20}(\bar D)$ may not be known since input depth map $D$ is unknown. Therefore there is a question of how to compute the conditioning embedding at inference time. Values of $\mathbf{Q}_{20}(\bar D)$ between $.7-1$. work for most images and it can be chosen heuristically. For instance, for DTU we uniformly assume a value of $.7$, which seems to work well. Note that any value of $\mathbf{Q}_{20}(\bar D)$ is presumably possible; it is only when this value is incompatible with the desired SDS camera radius that distillation may fail, since the cameras may intersect the visible content.

\subsection{Dataloader}

One significant engineering component of our work is our design of a streaming dataloader for multiview data, built on top of WebDataset \citep{webdataset}. Each dataset is sharded and each shard consists of a sequential tar archive of scenes. The shards can be streamed in parallel via multiprocessing. As a shard is streamed, we yield random pairs of views from scenes according to a ``rate'' parameter that determines how densely to sample each scene. This parameter allows a trade-off between fully random sampling (lower rate) and biased sampling (higher rate) which can be tuned according to the available network bandwidth. Individual streams from each dataset are then combined and sampled randomly to yield the mixture dataset. We will release the code together with our main code release.

\section{Details: NeRF prediction and distillation}
\label{appendix:nerf}

\subsection{SDS Anchoring} We propose SDS anchoring in order to increase the diversity of synthesized scenes. We sample 2 anchors at 120 and 240 degrees of azimuth relative to the input camera. 

One potential issue with SDS anchoring is that if the samples are 3D-inconsistent, the resulting generations may look unusual. Furthermore, traditional SDS already performs quite well except if the criterion is diverse backgrounds. Therefore, to implement anchoring, we randomly choose with probability $.5$ either the input camera and view or the nearest sampled anchor camera and view as guidance. If the guidance is an anchor, we "gate" the gradients flowing back from SDS according to the depth of the NeRF render, so that only depths above a certain threshold ($1.0$ in our experiments) receive guidance from the anchors. This seems to mostly mitigate artifacts from 3D-inconsistency of foreground content, while still allowing for rich backgrounds. We show video results for SDS anchoring on the webpage.

\subsection{Hyperparameters} NeRF distillation via involves numerous hyperparameters such as for controlling lighting, shading, camera sampling, number of training steps, training at progressively increasing resolutions, loss weights, density blob initializations, optimizers, guidance weight, and more. We will share a few insights about choosing hyperparameters for scenes here, and release the full configs as part of our code release.

\noindent\paragraph{Noise scheduling:} We found that ending training with very low maximum noise levels such as $.025$ seemed to benefit results, particularly perceptual metrics like LPIPS. We additionaly found a significant benefit on 360-degree scenes such as in the Mip-NeRF 360 dataset to scheduling the noise "anisotropically;" that is, reducing the noise level more slowly on the opposite end from the input view. This seems to give the optimization more time to solve the challenging 180-degree views at higher noise levels before refining the predictions at low noise levels.

\noindent\paragraph{Miscellaneous:} Progressive azimuth and elevation sampling following \citep{magic123} was also found to be very important for training stability. Training resolution progresses stagewise, first with batch size 6 at 128x128 and then with batch size $1$ at $256\times 256$.

\section{Experimental setups}
\label{appendix:setups}

For our main results on DTU and Mip-NeRF 360, we train our model and Zero-1-to-3 for $60,000$ steps. Performance for our method seems to saturate earlier than for Zero-1-to-3, which trained for about $100,000$ steps; this may be due to the larger dataset size. Objaverse, with $800,000$ scenes, is much larger than the combination of RealEstate10K, ACID, and CO3D, which are only about $95,000$ scenes in total.

For the retrained PixelNeRF baseline, we retrained it on our mixture dataset of CO3D, ACID, and RealEstate10K for about $560,000$ steps.

\subsection{Main results}

For all single-image NeRF distillation results, we assume the camera elevation, field of view, and content scale are given. These parameters are identical for all DTU scenes but vary across the Mip-NeRF 360 dataset. For DTU, we use the standard input views and test split from from prior work. We select Mip-NeRF 360 input view indices manually based on two criteria. First, the views are well-approximated by a 3DoF pose representation in the sense of geodesic distance between rotations. This is to ensure fair comparison with Zero-1-to-3, and for compatibility with Threestudio's SDS sampling scheme, which also uses 3 degrees of freedom. Second, as much of the scene content as possible must be visible in the view. The exact values of the input view indices are given in Table \ref{tab:mipnerf_input_views}.

The field of view is obtained via COLMAP. The camera elevation is set automatically via computing the angle between the forward axis of the camera and the world's $XY$-plane, after the cameras have been standardized via PCA following \citet{mipnerf360}. 

One challenge is that for both the Mip-NeRF 360 and DTU datasets, the scene scales are not known by the zero-shot methods, namely Zero-1-to-3, our method, and our retrained PixelNeRF. Therefore, for the zero-shot methods, we manually grid search for the optimal world scale in intervals of $.1$ to find the appropriate world scale for each scene in order to align the predictions to the generated scenes. Between five to nine samples within $[.3, .4, .5, .6, .7, .8, .9, 1., 1.1, 1.2, 1.3, 1.4, 1.5]$ generally suffices to find the appropriate scale. Even correcting for the scale misalignment issue in this way, the zero-shot methods generally do worse on pixel-aligned metrics like SSIM and PSNR compared with methods that have been fine-tuned on DTU.

\subsection{User study} We conduct a user study on the seven Mip-NeRF 360 scenes, comparing our method with and without SDS anchoring. We received 21 respondents. For each scene, respondents were shown 360-degree novel view videos of the scene inferred both with and without SDS anchoring. The videos were shown in a random order and respondents were unaware which video corresponded to the use of SDS anchoring. Respondents were asked:

\begin{enumerate}
    \item Which scene seems more realistic?
    \item Which scene seems more creative?
    \item Which scene do you prefer?
\end{enumerate}   

Respondents generally preferred the scenes produced by SDS anchoring, especially with respect to ``Which scene seems more creative?''

\subsection{Ablation studies}

We perform ablation studies on dataset selection and camera representations. For 2D novel view synthesis metrics, we compute metrics on a held-out subset of scenes from the respective datasets, randomly sampling pairs of input and target novel views from each scene. For 3D SDS distillation and novel view synthesis, our settings are identical to the NeRF distillation settings for our main results except that we use shorter-trained diffusion models. We train them for 25,000 steps as opposed to 60,000 steps for computational constraint reasons.

\begin{table}
\centering
\begin{tabular}{ccc}
\toprule
    Scene name & Input view index & Content scale \\
    \midrule
    bicycle &  98 & .9 \\
    bonsai & 204 & .9 \\
    counter & 95 & .9 \\
    garden & 63 & .9 \\
    kitchen & 65 & .9 \\
    room & 151 & 2. \\
    stump & 34 & .9 \\
    \bottomrule
\end{tabular}
\caption{Setup for the Mip-NeRF 360 dataset}.
\label{tab:mipnerf_input_views}
\end{table}

\end{document}